\documentclass[11pt]{article}

\usepackage[british]{babel}

\usepackage{acl}

\usepackage{hyperref}

\usepackage{times}
\usepackage{latexsym}
\usepackage{microtype}
\usepackage{inconsolata}

\usepackage{arydshln}
\usepackage[T1]{fontenc}
\usepackage[utf8]{inputenc}

\usepackage{flafter}
\usepackage[section]{placeins}

\usepackage{graphicx}
\usepackage{subcaption}
\usepackage{float}

\usepackage{array}
\usepackage{multirow}
\usepackage{adjustbox}

\usepackage{url}
\usepackage{amsmath}
\usepackage{amssymb}
\usepackage{mathtools}

\usepackage{caption}

\usepackage{subcaption}
\usepackage{tcolorbox}
\usepackage{xspace}
\usepackage{enumitem}

\usepackage{array}
\usepackage{makecell}
\usepackage{colortbl}
\usepackage{boldline}
\usepackage{multirow}
\usepackage{xcolor}
\usepackage{tabularx}

\usepackage{tikz}
\usetikzlibrary{calc}

\newcolumntype{M}[1]{>{\centering\arraybackslash}m{#1}}

\title{Measuring Embedding Sensitivity to Authorial Style in French:\\ Comparing Literary Texts with Language Model Rewritings}

\author{
  {\bf Benjamin Icard}$^{1}$ \quad
  {\bf Lila Sainero}$^{1}$ \quad
  {\bf Alice Breton}$^{1}$ \\
  {\bf Evangelia Zve}$^{1,2}$ \quad
  {\bf Jean-Gabriel Ganascia}$^{1}$ \vspace{0.1in} \\
  $^{1}$ LIP6, Sorbonne University, CNRS, France \\
  $^{2}$ Infopro Digital, France
}

\begin{document}

\maketitle

\begin{abstract} 

Large language models (LLMs) can convincingly imitate human writing styles, yet it remains unclear how much stylistic information is encoded in embeddings from any language model and retained after LLM rewriting. We investigate these questions in French, using a controlled literary dataset to quantify the effect of stylistic variation via changes in embedding dispersion. We observe that embeddings reliably capture authorial stylistic features and that these signals persist after rewriting, while also exhibiting LLM-specific patterns. These analytical results offer promising directions for authorship imitation detection in the era of language models.

%Large language models (LLMs) can convincingly imitate human writing styles, including in lower-resource languages such as French. However, it remains unclear how much stylistic information is encoded in embeddings produced by LLMs and other language models. We examine this question in French using a literary dataset designed to separate style from content in embedding space. Results show that embeddings capture style in both settings, but the patterns differ in ways that matter for computational stylometry. Our findings support more style-aware evaluation of text embeddings for LLM-generated text, especially for languages with lower-resources than English. 

%Large language models (LLMs) can convincingly imitate human writing style, even in lower-resource languages such as French. Yet it remains unclear how much writing style is captured by text embeddings, beyond semantic content, and how this changes after LLM rewriting. This paper proposes to study this question in French, assembling a controlled dataset of authorial literary excerpts and LLM stylistic imitations of them with topic held constant. Across thirteen embedding models, we measure sensitivity to authorial stylistic features using dispersion-based metrics. We find measurable stylistic signals in both human and rewritten texts, but with preservation varying across rewriting models and target author complexity. These findings call for more style-aware evaluation of text embeddings in human and LLM-generated settings.
 %The code, experimental data and analytical results are available at: \url{https://anonymous.4open.science/r/style-embedding-sensitivity-BD96}.

\end{abstract}

\section{Introduction}

Computational stylometry has long been central to digital humanities, using natural language processing (NLP) for authorship attribution and stylistic comparison \cite{stamatatos2009survey,koppel2011wild}. This has involved quantifying inter-textual distance and developing feature-based methods to support author verification \cite{savoy2012comparative,savoy2012specificvocab,cafiero2019moliere}.

In recent years, embedding methods, particularly Transformer-based contextual models \citep{vaswani2017attention,devlin2018bert}, have improved literary authorship attribution through richer textual representations of authorial profiles \citep{terreau2021writing,kim-etal-2025-leveraging}. Yet embedding-based analysis often centers on semantic tasks, especially topic modeling \citep{bianchi2021pre} and content similarity \citep{rockmore2025literary}, rather than style in embeddings, aside from a few studies \citep{wegmann2022author,icard2025embedding}. Assessing more systematically how embeddings capture authorial style would improve the explainability of embedding representations and strengthen authorship characterization and attribution in computational stylometry.

The increasing capabilities of large language models (LLMs) to imitate human-authored styles make this issue more urgent. Recent work on style transfer shows that LLMs can reproduce salient stylistic features in literary settings \citep{mikros2025beyond}, but also that they still struggle to imitate more implicit writing styles in extra-literary settings \citep{wang2025catch}. Because LLMs are language models, this reinforces the need to assess the sensitivity of embedding vectors to style, and the extent to which LLMs preserve or alter that sensitivity during style transfer \citep{huang2025authorship}.

This paper reports a controlled experiment on French literary texts evaluating embedding sensitivity to target stylistic features in human-authored texts and their LLM imitations. We compile a dataset of French excerpts from Tufféry \cite{tuffery2000style}, Proust \cite{proust1913tome1}, Céline \cite{celine1932}, and Yourcenar \cite{yourcenar1951}, together with \mbox{GPT-,} Mistral-, and Gemini-based stylistic imitations under fixed-topic conditions. We encode all texts with thirteen embedding models and quantify aggregated stylistic effects on embedding dispersion across human authors and imitated texts.

Section~\ref{sec:related} reviews prior work on embedding sensitivity to writing style, with particular attention to the human–LLM distinction. Section~\ref{sec:corpus} introduces our French literary dataset and the evaluation of style transfer under LLM generation. Section~\ref{sec:experiments} presents the vectorization of the dataset with thirteen embedding models, a structural evaluation with clustering, and results on how writing style affect embedding dispersion in human-authored texts and LLM imitations. Section~\ref{sec:discussion} examines LLM-specific effects on embedding sensitivity, 
bringing explainability to the style transfer evaluation used to validate the dataset. Finally, Section~\ref{sec:conclusion} concludes and outlines directions for future work. 

All reproducibility materials and results are available at: \url{https://github.com/sma-libra/style-embedding-sensitivity}

\section{Related Work}
\label{sec:related}

\textbf{Feature-Based Stylometry.} Computational stylometry classically relies on surface features such as word frequencies, character $n$-grams \citep{cavnar1994n,rios2022detection} and punctuation patterns \citep{faye2024exposing}, often represented with TF-IDF term weighting \citep{salton1988term,bui2011writer}, to capture authorial voice \citep{verma2019lexical,herrmann2021computational,mani2022computational}. 

\paragraph{Style Embeddings.} More recent work examines whether embedding-based representations also encode feature-based stylistic information \citep{liu2024team}. \citet{terreau2021writing} propose an author-verification framework that tests whether embedding spaces encode stylistic features rather than mainly semantic content. Using this framework, they show that specialized Doc2Vec-based author embeddings \citep{le2014distributed,ganesh2016author2vec,maharjan-etal-2019-jointly} are often more semantically driven, while simpler pretrained sentence encoders such as USE-DAN \citep{cer2018universal} and SBERT \citep{reimers2019sentence} can perform better on stylistic feature families. %(letters, numbers, structural, punctuation, function words, POS tags, NER, and complexity/readability indexes).

\paragraph{Measuring Sensitivity.} A related line of work addresses style-semantics conflation more directly through topic-controlled or style-sensitive embedding representations. More specifically, \citet{wegmann2022author} show that content-controlled training improves style-topic separation in BERT-based representations. Adjacent work by \citet{chen2023writing} shows that style-sensitive encoders can support efficient detection of stylistic shifts in multi-author documents. In parallel, \citet{patel2023learning} introduce the style embedding model LISA whose dimensions correspond to stylistic features. Most directly related to our approach, \citet{icard2025embedding} use Queneau’s fixed-topic variations to measure embedding sensitivity to style under rewriting by a single LLM, rather than focusing specifically on author-defining features and their preservation.

\paragraph{LLM Style Transfer.} Recent research on LLMs’ representation and high-fidelity transfer of literary style is motivated by their demonstrated capacity to imitate authorial writing. In literary contexts, \citet{mikros2025beyond} show that GPT-4o reproduces salient authorial stylistic features under thematic control, while \citet{sarfati-etal-2025-whats} and \citet{hicke2025innermusic} show that literary style is reflected in LLM internal representations. In parallel, \citet{huang-etal-2024-large} study LLMs for authorship analysis, while \citet{horvitz-etal-2024-tinystyler} propose TinyStyler, a lightweight approach to few-shot style transfer conditioned on author embeddings.

Taken together, these works have examined embedding sensitivity to style in smaller language models and, more recently, in LLMs. However, little work has investigated whether embeddings are \textit{reliably} sensitive to the defining stylistic features of specific authors, and to what extent this sensitivity remains measurable \textit{after} LLM rewriting. We investigate this question in French using stylistically marked and diverse literary texts.

%Building on these advances, we make two contributions. First, we test whether embedding sensitivity to style aligns with established authorial features from stylometric literature, extending beyond prior work that primarily measures sensitivity without validating against known stylistic markers. Second, unlike previous studies, which analyze human texts only or use single-model generation, we assess whether stylistic signatures remain detectable after rewriting by three different LLM architectures.

%A neighboring line of research detects machine-generated text using statistical or behavioral cues rather than stylistic interpretation. Approaches include perplexity thresholds and perturbation-based likelihood curvature \citep{gehrmann2019gltr,ippolito2020automatic,mitchell2023detectgpt,bao2024fast}, paired-model and ensemble detectors \citep{hans2401spotting,dubois-etal-2025-mosaic}, rewriting stability \citep{mao2024raidar}, and watermarking \citep{kirchenbauer2023watermark,zhang2023robust}. Classifiers fine-tuned on RoBERTa or T5 perform well in-domain but collapse across unseen generators \citep{zellers2019defending,guo2023close,liu2024benchmark,conneau2020unsupervised,raffel2020exploring}. Robustness filters further mitigate adversarial risk \citep{kuznetsov-etal-2024-robust}. These strategies achieve partial success but treat style only implicitly. By contrast, we ask whether \emph{stylistic signals in embedding space} persist under generative rewriting, offering complementary evidence for distinguishing human from machine-authored text.

\section{Dataset}
\label{sec:corpus}

\subsection{Text Materials}

We assembled a dataset of 1,248 French literary texts that combines human-authored originals with LLM-generated imitations that reproduce the human authors' styles under fixed-topic conditions.  

\begin{figure*}[t]
    \centering
    
    % Optional: adjust padding between image and border
    \setlength{\fboxsep}{0pt} % no padding
    \setlength{\fboxrule}{0.3pt} % thickness of border
    
    \fbox{\includegraphics[width=\linewidth]{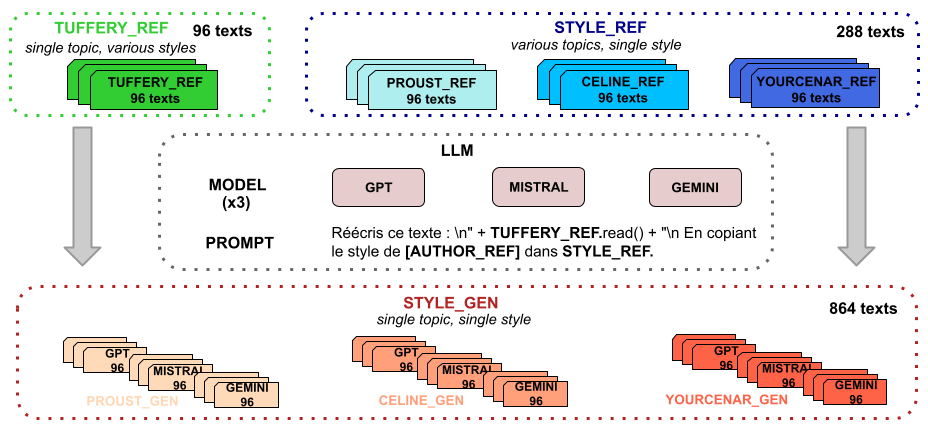}}

  \caption{Textual corpora and generation scheme used to construct the \textsc{StyloGen} embedding dataset. The reference corpora are \textsc{Tuffery\_ref} (single topic, various styles) and \textsc{Style\_ref} (various topics, single style per author). Three LLMs generate \textsc{Style\_gen} by rewriting Tufféry texts in the styles of the authors in \textsc{Style\_ref}, using the French prompt shown.}
    \label{fig:overview}
\end{figure*}

\subsubsection{Reference Corpus}
\label{ssec:refcorp}

We first compiled a \textit{reference corpus} of 384 French original literary texts, separated in two groups according to topic and style variation.

The first group, \textsc{Tuffery\_ref}, consists of 96 texts extracted from Stéphane Tufféry's \textit{Le style mode d'emploi} \cite{tuffery2000style}.\footnote{We excluded 3 of the 99 texts from Tufféry’s \textit{Le style mode d'emploi} when constructing $\textsc{Tuffery\_ref}$ because they departed from the bus-journey topic.} We use this collection because, like Raymond Queneau’s \textit{Exercices de style} \citep{queneau1947exercices} used in \citet{icard2025embedding}, it keeps a single topic across all texts: the story of a bus journey in Paris. But unlike Queneau’s variations, however, Tufféry structures stylistic variation more systematically, including explicit pastiches of major French authors such as Balzac, Flaubert, and Hugo. \textsc{Tuffery\_ref} therefore provides a fixed-topic, style-diverse baseline for stylometric comparison.

The second group, \textsc{Style\_ref}, comprises 288 texts of length comparable to $\textsc{Tuffery\_ref}$, evenly distributed across three subclasses:%\footnote{The text-length distribution in $\textsc{Tuffery\_ref}$ was used to construct $\textsc{Proust\_ref}$, $\textsc{Celine\_ref}$, and $\textsc{Yourcenar\_ref}$ by selecting texts of comparable length.}

\begin{itemize}
\vspace{-0.2em}
\item \textsc{Proust\_ref}: 96 texts from Proust’s \textit{Du côté de chez Swann}, the first volume of \textit{À la recherche du temps perdu} \cite{proust1913tome1};

\vspace{-0.2em}
\item \textsc{Celine\_ref}: 96 texts from Céline’s \textit{Voyage au bout de la nuit} \cite{celine1932};

\vspace{-0.2em}
\item \textsc{Yourcenar\_ref}: 96 texts from Yourcenar’s \textit{Mémoires d’Hadrien} \cite{yourcenar1951}.
\end{itemize}

The authors were selected to maximize stylistic contrast across four central dimensions in French literature \citep{MagriMourgues2006CorpusStylistique,Brunet2014Lexicometrie,Hough2016OxfordHandbookNames}: structural complexity, morphosyntactic richness, lexical diversity, and referential anchoring. Proust exemplifies structural complexity (long sentences, high average word length) and morphosyntactic richness (diverse POS patterns), with recurrent proper-name allusions (named entities) \citep{brunet1982style}.
Céline foregrounds a spoken register (short sentences and words, high pronoun/verb share) alongside lexical diversity and nonstandard forms (higher entropy, distinctive character distributions) \citep{Zabojnikova2006CelineOralPopulaire,Styhre2011CelineHyperbole,SilvestreDeSacy2025Hypersegmentation}.
Yourcenar sustains balanced prose densely anchored in historical and geographic reference (named-entity density) \citep{Colvin2005BaroqueFictions,Broche2022YourcenarHistoriens}.
%Table~\ref{tab:author_styles_nlp} provides an overview of their widely acknowledged stylistic characteristics and the corresponding stylometric features for measurement.

%To do this, for each of the three novels in \textsc{Style\_ref}, we manually divided the opening section into 96 texts of approximately equal length. We selected division points at paragraph breaks whenever possible, or otherwise at sentence boundaries that maintained the coherence of the text.
%\textcolor{blue}{Here we should be a bit more precise about how we proceeded to obtain that}

\begin{table*}[t]
\centering
\small
\resizebox{\linewidth}{!}{%
\begin{tabular}{llccc||c}
\hline
\multicolumn{2}{l}{\textbf{Validation Set}}  &
\textsc{Proust\_ref} & \textsc{Celine\_ref} & \textsc{Yourcenar\_ref} & \textsc{Style\_ref} (held-out 20\%) \\ \hline
\multirow{2}{*}{\shortstack[l]{\\Corpus-Level}} & Macro-F1             & --- & --- &  --- & 0.965 \\
& Accuracy    & 0.947 & 0.947 & 1.000 & \textbf{0.966} \\

\hline
\end{tabular}}
\vspace{1em}

\resizebox{\linewidth}{!}{%
\begin{tabular}{llccc||c}
\hline
\multicolumn{2}{l}{\textbf{Test Set}} &
\textsc{Proust\_gen} & \textsc{Celine\_gen} & \textsc{Yourcenar\_gen} & \textsc{Style\_gen} (100\%) \\ \hline

\multirow{2}{*}{\shortstack[l]{Corpus-Level}} & Macro-F1 & --- & --- &  --- & 0.669 \\
& Transfer Accuracy  & 0.601 & 0.722 & 0.670 & \textbf{0.664} \\ \hline

\multirow{3}{*}{\shortstack[l]{Per-Class}}
&  \quad \textit{when \textsc{GPT} is used}     & 0.490 & 0.542 & 0.646 & 0.559 \\ %\cline{2-6}
& \quad \textit{when \textsc{Mistral} is used} & \textcolor{blue}{0.844} & \textcolor{blue}{0.812} & 0.625 & \textcolor{blue}{0.760} \\ %\cline{2-6}
& \quad \textit{when \textsc{Gemini} is used}  & 0.469 & \textcolor{blue}{0.812} & \textcolor{blue}{0.740} & 0.674 \\ 
\hline
\end{tabular}}
\caption{Style transfer results with TF-IDF character $3$-$5$-grams + \texttt{LinearSVC}. We report corpus-level performances on \textsc{Style\_ref} (held-out 20\%) and \textsc{Style\_gen} (100\%), and per-class transfer accuracy for each target author label. 
We indicate corpus-level accuracy in bold, and the highest per-class transfer accuracy across LLMs in blue.}
\label{tab:style_transfer_validator}
\end{table*}

\subsection{Generated Corpus}

We constructed the generated corpus, \textsc{Style\_gen}, by prompting LLMs to rewrite the 96 texts from $\textsc{Tuffery\_ref}$ in the writing styles of $\textsc{Proust\_ref}$, $\textsc{Celine\_ref}$, and $\textsc{Yourcenar\_ref}$, respectively. To ensure both diversity of behavior and architectural variety, we employed three LLMs for each target style: \href{https://platform.openai.com/docs/models/gpt-4o}{\texttt{GPT-4o}}, \href{https://docs.mistral.ai/getting-started/models/models_overview/}{\texttt{mistral-large-2411}}, and \href{https://ai.google.dev/gemini-api/docs/models#gemini-1.5-flash}{\texttt{gemini-1.5-flash}}. For simplicity's sake, we now refer to these LLM versions as \textsc{Gpt}, \textsc{Mistral}, and \textsc{Gemini}.

The group \textsc{Style\_gen} comprises 864 texts evenly distributed across three subclasses:

\begin{itemize}
\item \textsc{Proust\_gen}: 288 texts produced by prompting each of the three generative models to rewrite Tufféry's texts in the style of Proust’s \textit{Du côté de chez Swann};
\item \textsc{Celine\_gen}: 288 texts produced by prompting each of the three generative models to rewrite Tufféry's texts in the style of Céline’s \textit{Voyage au bout de la nuit};
\item \textsc{Yourcenar\_gen}: 288 texts produced by prompting each of the three generative models to rewrite Tufféry's texts in the style of Yourcenar’s \textit{Mémoires d’Hadrien}.
\end{itemize}

Figure~\ref{fig:overview} presents an overview of the pipeline followed to assemble our textual corpora, including the French prompt used. The English translation of the prompt is: \textit{``Rewrite this text: \textbackslash n" + \textbf{\textsc{Tuffery\_ref}}.read() + ``\textbackslash n By copying the style of \textbf{\textnormal{[}\textsc{Author\_ref}\textnormal{]}} into \textbf{\textsc{Style\_ref}}.''}

\subsection{Style Transfer Evaluation}
\label{ssec:style_transfer_eval} 

We assessed style transfer from \textsc{Style\_ref} to \textsc{Style\_gen} as a function of the LLM used as the generator. Specifically, we evaluated the extent to which texts in \textsc{Proust\_gen}, \textsc{Celine\_gen}, and \textsc{Yourcenar\_gen} were classified as \textsc{Proust\_ref}, \textsc{Celine\_ref}, and \textsc{Yourcenar\_ref}, respectively. 

\paragraph{Character $n$-Gram Validator.}
For this task, we used a validator that represents each document with TF-IDF-weighted character $3$-$5$-grams and classifies it with a linear support vector classifier \citep{cortes1995support}, implemented with \texttt{LinearSVC} from scikit-learn, into one of three author-style labels: \textsc{Proust\_ref}, \textsc{Celine\_ref}, \textsc{Yourcenar\_ref}. We chose this validator because character $n$-grams are strong surface-level features of authorial style \citep{kevselj2003n,stamatatos2009survey}, and because it is embedding-independent, avoiding confounds in our later sensitivity analyses.\footnote{Appendix \ref{ssec:styletransfunct} (Table~\ref{tab:style_transfer_validator_function}) reports additional style transfer results on the corpus using another linear \texttt{LinearSVC} classifier, this time with function-word frequencies as a classical stylometric baseline for authorial style \citep{burrows2002delta}.}

Methodologically, we split \textsc{Style\_ref} into train/validation sets and use \textsc{Style\_gen} as the test set:
\begin{itemize}
\item[—] \textbf{Train/Validation:} using \textsc{Style\_ref} only (i.e., \textsc{Proust\_ref}, \textsc{Celine\_ref}, and \textsc{Yourcenar\_ref}) split into 80\% for train and 20\% for validation (stratified by author, with seed \(=42\));
\item[—] \textbf{Test:} using 100\% of \textsc{Style\_gen} (i.e., \textsc{Proust\_gen}, \textsc{Celine\_gen}, and \textsc{Yourcenar\_gen}), evaluated without adaptation to measure style transfer from human to LLM-generated text.
\end{itemize}

Table~\ref{tab:style_transfer_validator} reports the validator's performances on the held-out \textsc{Style\_ref} corpus (20\%) and its \textit{transfer} accuracy on the \textsc{Style\_gen} corpus (100\%), broken down by imitated author class and by LLM used as a generator.

At the corpus level, the validator achieves near-ceiling accuracy on held-out human texts from \textsc{Style\_ref} (Accuracy: \(0.947\) to \(1.00\)). When applied to \textsc{Style\_gen}, transfer accuracy drops substantially, but remains well above the three-way chance level (\(1/3\)) (Accuracy: \(0.664\)).

Transfer accuracy on generated texts varies with the LLM used for imitation. \textsc{Proust\_gen} is best recognized when generated by \textsc{Mistral} ($0.844$), \textsc{Celine\_gen} is best recognized when generated by \textsc{Mistral} or \textsc{Gemini} ($0.812$ in both cases), and \textsc{Yourcenar\_gen} is best recognized when generated by \textsc{Gemini} ($0.740$). By contrast, \textsc{GPT} yields the lowest accuracies across the three generated corpora. The confusion matrix complementing these results is provided in Appendix~\ref{ssec:appendix_confusions} (Figure~\ref{fig:cm_char_svc}).

%Taken together, these results indicate that the generated texts preserve substantial target-style features detectable from surface-form features in authorial texts, while not implying full stylistic fidelity. 

Beyond style transfer validation from \textsc{Style\_ref} to \textsc{Style\_gen} across LLM generators, we now analyze embedding representations of the same full textual corpora to compare embedding sensitivity to stylistic features before and after LLM rewriting.

%We now present embedding representations of the full dataset across multiple models, as a basis for a feature-based analysis of embedding sensitivity to writing style in the human-authored corpus and after LLM imitation.

\section{Embedding Sensitivity to  Writing Style}
\label{sec:experiments}

\subsection{Dataset Vectorization}

\paragraph{Embedding Model Selection.} The reference corpora \textsc{Tuffery\_ref} and \textsc{Style\_ref}, along with the generated corpus \textsc{Style\_gen}, were embedded using thirteen models (Table~\ref{tab:meanumap}), with their embedding dimensionalities reported. Selection criteria were architectural and dimensional diversity, computational efficiency, and strong performance on the Massive Text Embedding Benchmark (MTEB; November 2025 version)\footnote{\url{https://huggingface.co/spaces/mteb/leaderboard}} \citep{muennighoff2022mteb}. The resulting embedding dataset for all thirteen model is available in our data repository. 

%\footnote{See Section Ethical Considerations for details.} 

\paragraph{Clustering Validation.} To assess embedding space separability across the three corpora (\textsc{Tuffery\_ref}, \textsc{Style\_ref}, \textsc{Style\_gen}), we run k-means clustering~\cite{hartigan1979algorithm} on the full-dimensional embeddings of the thirteen models, setting \(k=3\).\footnote{Using \textit{scikit-learn} \texttt{KMeans} with default parameters and \texttt{random\_state}=0.} Clustering quality is evaluated using purity~\citep{manning2008introduction}, an external score in \([0,1]\) computed by comparing the induced clusters to the three ground-truth corpus labels~\citep{soni2024clutching}. Table~\ref{tab:meanumap} reports, for each embedding model, the purity score computed in the model’s full-dimensional space (FullD).

\begin{table}[h]
\centering
\tiny
\resizebox{\linewidth}{!}{%
\begin{tabular}{|l|c|c|}
\hline
\textbf{Embedding Model} & \textbf{FullD} & \textbf{Purity} \\
\hline
\href{https://huggingface.co/FacebookAI/xlm-roberta-large}{\texttt{xlm-roberta-large}} & 1024 & 0.7654 \\
\hline
\href{https://huggingface.co/intfloat/multilingual-e5-large}{\texttt{multilingual-e5-large}} & 1024 & 0.6836 \\
\hline
\href{https://docs.mistral.ai/capabilities/embeddings/}{\texttt{mistral-embed}} & 1024 & 0.6809 \\
\hline
\href{https://huggingface.co/intfloat/e5-base-v2}{\texttt{e5-base-v2}} & 768 & 0.6802 \\
\hline
\href{https://huggingface.co/distilbert/distilbert-base-uncased}{\texttt{distilbert-base-uncased}} & 768 & 0.6701 \\
\hline
\href{https://platform.openai.com/docs/guides/embeddings}{\texttt{text-embedding-3-small}} & 1536 & 0.6663 \\
\hline
\href{https://cloud.google.com/vertex-ai/generative-ai/docs/embeddings/get-text-embeddings}{\texttt{text-embedding-004}} & 768 & 0.6539 \\
\hline
\href{https://huggingface.co/OrdalieTech/solon-embeddings-large-0.1}{\texttt{solon-embeddings-large-0.1}} & 1024 & 0.6524 \\

\hline
\href{https://docs.voyageai.com/docs/embeddings}{\texttt{voyage-2}} & 1024 & 0.6493 \\

\hline
\href{https://huggingface.co/dangvantuan/sentence-camembert-base}{\texttt{sentence-camembert-base}} & 768 & 0.6408 \\
\hline
\href{https://huggingface.co/sentence-transformers/all-roberta-large-v1}{\texttt{all-roberta-large-v1}} & 1024 & 0.6242 \\

\hline
\href{https://huggingface.co/sentence-transformers/paraphrase-multilingual-mpnet-base-v2}{\texttt{paraphrase-multilingual-mpnet-base-v2}} & 768 & 0.6080 \\
\hline
\href{https://huggingface.co/sentence-transformers/all-MiniLM-L12-v2}{\texttt{all-MiniLM-L12-v2}} & 384 & 0.5721 \\

%\hline
%\textbf{Mean} & \textbf{916} & \textbf{0.6575} \\
%\hline
%\textbf{Median} & \textbf{1024} & \textbf{0.6539} \\

\hline
\end{tabular}}
\caption{Embedding dimensionality and k-means cluster purity (\(k=3\)) for each model in full-dimensional space (FullD), computed with respect to the three corpus labels (\textsc{Tuffery\_ref}, \textsc{Style\_ref}, \textsc{Style\_gen}). Models are ordered by purity scores.}
\label{tab:meanumap}

\end{table}

Recovery of the three corpus groups is reasonable across the thirteen embedding models, with mean purity ($0.6575$) and a close median ($0.6539$). At the level of individual models, \texttt{xlm-roberta-large} scores highest ($0.7654$) and \texttt{all-MiniLM-L12-v2} lowest ($0.5721$), with intermediate models spread smoothly across this range.

\paragraph{UMAP Reduction.} Beyond the consistent purity scores across models, Table~\ref{tab:meanumap} shows substantial variation in embedding dimensionality (from 384 to 1536). To enable comparisons across model-specific full-dimensional spaces (FullD), we apply Uniform Manifold Approximation and Projection (UMAP)~\cite{mcinnes2018umapjoss}. For each target dimensionality, we repeat the projection thirty times with different random seeds to account for UMAP's stochasticity.

\paragraph{Focus on 2D UMAP.} We tested UMAP reductions, specifically 2D, 3D, and 10D, to find the optimal sufficient dimension with respect to numerical fidelity to the FullD purity scores. To quantify alignment, we computed the mean absolute error (MAE) to FullD, as well as the maximum absolute error (MaxAE). 

\begin{figure}[h]
    %\centering
    %\begin{subfigure}[t]{0.48\linewidth}
        %\centering
        %\includegraphics[width=\linewidth]{Figures/image-18.png}
        %\caption{Correspondence at the level of \textsc{Tuffery\_ref}, \textsc{Style\_ref}, \textsc{Style\_gen}.}
        %\caption{}
        %\label{fig:dispplot_main}
    %\end{subfigure}
    %\hspace{0.015\linewidth}
    %\begin{subfigure}[t]{0.48\linewidth}
        \centering
        \includegraphics[width=\linewidth]{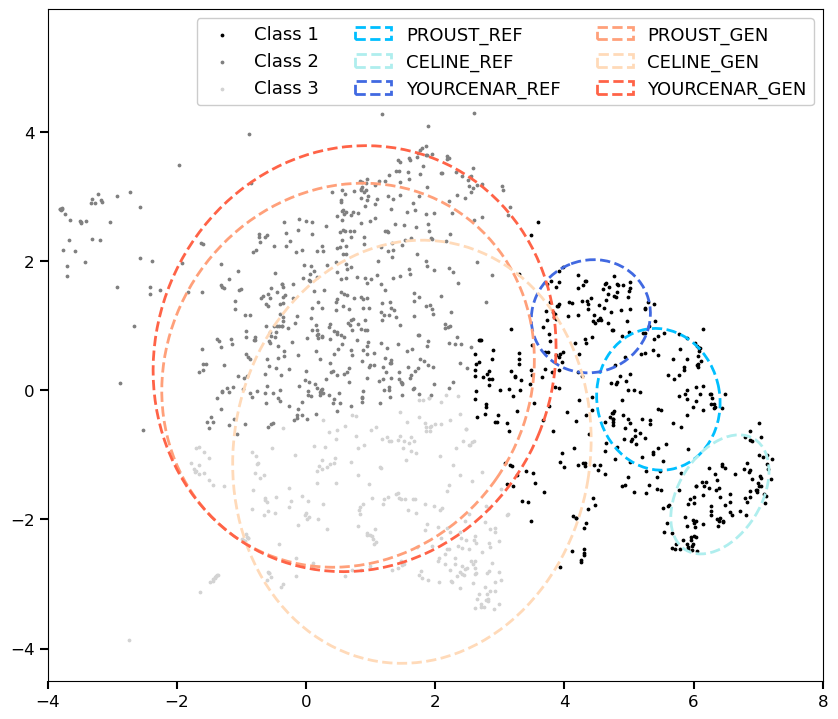}
        %\caption{Correspondence at the level of subclasses within \textsc{Style\_ref} and within \textsc{Style\_gen}.}
        %\caption{}
        \label{fig:dispplot_sub}
    %\end{subfigure}

\caption{2D UMAP projection of \texttt{xlm-roberta-large} embeddings on the dataset. Points indicate the three k-means clusters (Class 1-3), while dashed ellipses (visual guides, not k-means clusters) indicate label-based coverage regions for the human and generated corpora, drawn so that exactly \(80\%\) of the corpus lie inside the ellipse zone.}
    \label{fig:clusteringvalid}
\end{figure}

Among the UMAP reductions, we observed that 2D performed best, with the lowest MAE ($0.025$ for 2D, vs $0.036$ for 3D and $0.029$ for 10D) but also the smallest MaxAE ($0.056$ for 2D, vs $0.090$ for 3D and $0.066$ for 10D), making it the reduction best aligned with FullD purity. To visualize corpus-group structure, Figure~\ref{fig:clusteringvalid} shows the 2D UMAP projection of the \texttt{xlm-roberta-large} embedding model, which, as in FullD, obtains the best clustering purity score ($0.7627$).

%In 2D UMAP, as in FullD, \texttt{xlm-roberta-large} still achieves the highest purity ($0.7627$) and \texttt{all-roberta-large-v1} the lowest ($0.6192$), with \texttt{all-MiniLM-L12-v2} just slightly above ($0.6204$). 

%Figure \ref{fig:clusteringvalid} shows the 2D UMAP projections of clusters obtained with $\texttt{xlm-roberta-large}$ and their correspondence to the main classes and subclasses of \textsc{StyloGen}.

%\textsc{StyloGen}, with clear separation of the groups and strong correspondence with original labels, consistent with the purity score reported in Table~\ref{tab:meanumap} (0.7627). The marked separation between \textsc{Tuffery\_ref} and \textsc{Style\_ref} reflects both topical and stylistic differences, while the partial overlap between \textsc{Tuffery\_ref} and \textsc{Style\_gen} reflects topical alignment despite stylistic divergence.

The dashed ellipses in Figure~\ref{fig:clusteringvalid} suggest possible stylistic effects on the embedding representations. For \textsc{Proust\_ref}, \textsc{Celine\_ref}, and \textsc{Yourcenar\_ref} (right side of the figure), the clear separation between ellipses is consistent with differences in both topic and style across the human corpora, whereas the strong within-ellipse cohesion cannot be explained by topic, since topic varies within each corpus, and may instead reflect the homogenizing effect of shared authorial style. For \textsc{Proust\_gen}, \textsc{Celine\_gen}, and \textsc{Yourcenar\_gen} (left side of the figure), the substantial overlap is consistent with intended topic alignment on \textsc{Tuffery\_ref}, but the overlap remains only partial, especially for \textsc{Celine\_gen}, suggesting that embeddings may retain residual stylistic differences after LLM rewriting.

%Figure \ref{fig:clusteringvalid} reveals three dense clusters within \textsc{Style\_ref}, where ..., and within \textsc{Style\_gen}, where ... \textsc{Proust\_ref}, \textsc{Celine\_ref}, and \textsc{Yourcenar\_ref} remain well separated, as expected given their distinct and uncontrolled topics, whereas \textsc{Proust\_gen}, \textsc{Celine\_gen}, and \textsc{Yourcenar\_gen} overlap more substantially because they are expected to share the same aligned topic, namely \textsc{Tuffery\_ref}. 

These observations motivate measuring whether embeddings capture stylistic features beyond topic under fixed-topic rewriting. We now examine embedding sensitivity to author-characteristic features in French, both before and after LLM imitation.

%Although \textsc{Tuffery\_ref} and \textsc{Style\_gen} show substantial overlap in Figure~\ref{fig:clusteringvalid}, the correspondence is not complete. In addition, the 2D projection does not visibly separate \textsc{Proust\_gen}, \textsc{Celine\_gen}, and \textsc{Yourcenar\_gen}, despite differences in intended rewriting style. %Taken together, these observations motivate testing whether embeddings capture factors beyond topic under fixed-topic rewriting, including stylistic features and LLM-specific artifacts.

\subsection{Sensitivity Evaluation Metrics}
\label{ssec:sensimet}

Following \citet{icard2025embedding}, we compute dispersion-based metrics on an aggregated 2D UMAP reduction derived from the thirteen embedding models listed in Table~\ref{tab:meanumap}.

% \paragraph{Embedding Dispersion.} To quantify embedding dispersion, we define, across iterated 2D UMAP projections, \(d_X^{(i,j)}\) as the Euclidean distance between the \(i\)-th embedding vector and the centroid \(c_X^{(j)}\) of class \(X\) at iteration \(j\):

% \begin{equation}\label{eq:euclid_iter}
% d_X^{(i,j)} = \|v_X^{(i,j)} - c_X^{(j)} \|
% \end{equation}

% where $v_X^{(i,j)}$ is the $i$-th embedding vector of class $X$ in the $j$-th iteration, and $\| \cdot \|$ is the Euclidean norm.

% To capture the spatial distribution of high-dimensional embeddings, we calculate the mean Euclidean distance from the centroid of each class across all iterations, written $\bar{d}_X(i)$:

% \begin{equation}\label{eq:avg_distance}
% \bar{d}_X(i) = \frac{1}{30} \sum_{j=1}^{30} d_X^{(i,j)}
% \end{equation}

\paragraph{Embedding Dispersion.} 

For the $j$-th UMAP iteration, we define $d_X^{(i,j)}$ as the Euclidean distance of the $i$-th embedding vector from the centroid $c_X^{(j)}$ of class $X$ as follows:
\begin{equation}\label{eq:euclid_iter}
d_X^{(i,j)} = \|v_X^{(i,j)} - c_X^{(j)} \|
\end{equation}
where $v_X^{(i,j)}$ is the $i$-th embedding vector of class $X$ in the $j$-th iteration and $\| \cdot \|$ is the Euclidean norm.

To capture the spatial dispersion of embeddings in the UMAP target space, we calculate the mean Euclidean distance from the centroid of each class $X$ across all 30 UMAP iterations and all embeddings $N$ in $X$, written $\bar{d}_X$:

\begin{equation}\label{eq:avg_distance}
\bar{d}_X
= \frac{1}{N} \sum_{i=1}^{N} \left( \frac{1}{30} \sum_{j=1}^{30} d_X^{(i,j)} \right)
\end{equation}

%\vspace{-0.17cm}

%where $N$ is the total number of embedding vectors in class $X$.

%where $\bar{d}_X(i)$ is the averaged Euclidean distance of the $i$-th embedding vector for class $X$.

%Finally, the overall mean distance $\bar{d}_X$ for class $X$ across all embeddings is given by:

%\begin{equation}\label{eq:centroid}
%\bar{d}_X = \frac{1}{N} \sum_{i=1}^{N} \bar{d}_X(i)
%\end{equation}

%\vspace{-0.17cm}
%where $N$ is the total number of embedding vectors in class $X$.

\begin{figure*}[!h]
    \centering
    \begin{subfigure}[t]{0.46\textwidth}
        \centering
        \includegraphics[width=\linewidth]{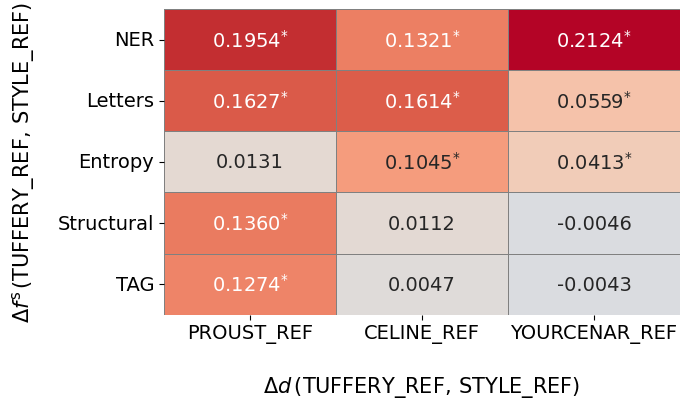}
       
        \phantomsubcaption\label{fig:corr_author_ref_a}
        \par\vspace{-0.2em} \makebox[\linewidth][c]{\hspace*{17mm}\small{(a)}}
        
        %\caption{Correlations ($r$) between embedding dispersion shifts ($\Delta d$) and stylistic feature shifts ($\Delta f^s$) for \textsc{Style\_ref} three corpora relative to \textsc{Tuffery\_ref}. Asterisks indicate $p<0.01$, with Bonferroni correction.}
        \label{fig:corr_author_ref}
    \end{subfigure}
    \hfill
    \begin{subfigure}[t]{0.46\textwidth}
        \centering
        \includegraphics[width=\linewidth]{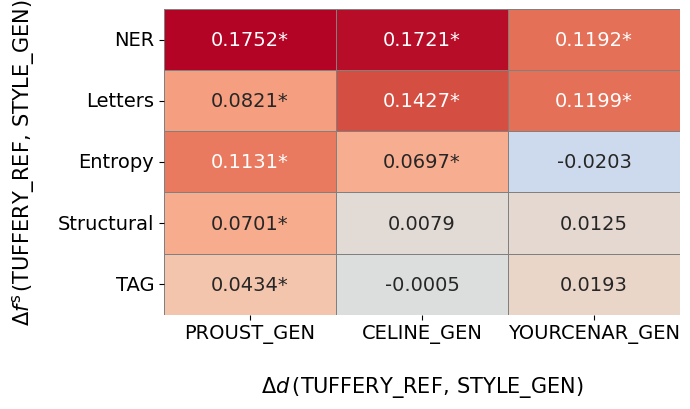}
        \phantomsubcaption\label{fig:corr_author_ref_b}
        \par\vspace{-1.3em} \makebox[\linewidth][c]{\hspace*{17mm}\small{(b)}}
        \label{fig:corr_author_gen}
    \end{subfigure}
    \vspace{-1.5em}
    \caption{Pearson correlations ($r$) between embedding dispersion shifts in 2D UMAP reduction ($\Delta d$) and stylistic feature shifts ($\Delta f^s$) for each author-labeled corpus, comparing \textsc{Tuffery\_ref} with (a) the three human-authored corpora in \textsc{Style\_ref}, and (b) the three style-imitated corpora in \textsc{Style\_gen}. Asterisks indicate $p<0.01$ after Bonferroni correction.}
        \label{fig:corr_author2D}
\end{figure*}

\vspace{-0.em}

%\begin{figure}[h]
%   \centering
%\includegraphics[width=\linewidth]{Figures/xml_roberta_large_disp.png}
%\caption{2D UMAP contour plots of embedding dispersion with $\texttt{xlm-roberta-large}$ on \textsc{StyloGen}. For each class, a dot indicates the centroid, the outer contour shows the overall spread around the centroid (last seed), isolines indicate density differences, and $\bar{d}_X$ denotes the mean centroid distance for class $X$.}
%\label{fig:disp}
%\end{figure}

\paragraph{Target Stylistic Features.} To analyze embedding sensitivity to stylistic variation in human texts and LLM rewritings, we must retain features that reflect the author-characteristic dimensions associated with Proust, Céline, and Yourcenar, as described in Section~\ref{ssec:refcorp}.

To this end, we drew on the stylometric framework proposed by \citet{terreau2021writing}, which is based on a comprehensive inventory of eight stylistic features families,\footnote{The complete list and description of the eight feature families are available at: \url{https://github.com/EnzoFleur/style_embedding_evaluation/}.} from which we retained a subset of five (\textit{structural features}, \textit{part-of-speech tags}, \textit{indexes of lexical diversity}, \textit{letter frequencies}, and \textit{named entities}) most relevant to authorial style along the core dimensions previously considered:

\begin{itemize}
    \item \textbf{Structural:} mean values of selected \textit{structural features}, specifically word length and sentence length, normalized by text length;  
    \item \textbf{Morphosyntax:} frequency of \textit{part-of-speech tags} (nouns, verbs, adjectives) to capture grammatical diversity patterns;  
    \item \textbf{Lexicon:} \textit{lexical diversity} (using Shannon Entropy, see \citealt{shannon1948mathematical}) and \textit{letter}-level patterns (character unigram and capitalization frequencies);
    \item \textbf{Referentiality:} density of \textit{named entities} (e.g., persons, locations, organizations) per sentence, as a proxy for referential content.  
\end{itemize}  

For convenience, we write \(f^{s}_X(i)\) to denote the average frequency value $f$ of the stylistic feature $s$ measured on the \(i\)-th document of class \(X\).

\paragraph{Dispersion-Style Correlations.}\hspace{-0.7em} In the following analyzes, \textsc{Tuffery\_ref} serves as the reference class to study interactions between embedding dispersion and writing style on the corpus. 

We denote by $\Delta d(\textsc{Tuffery\_ref}, Y)$ the difference in embedding dispersion between \textsc{Tuffery\_ref} and another class $Y$ (human or generated, here), and by $\Delta f^s(\textsc{Tuffery\_ref}, Y)$ the difference in frequency of stylistic feature $s$ between the two classes.
More formally:  
\begin{equation}\small
\Delta d (\textsc{Tuffery\_ref}, Y) = d_\textsc{Tuffery\_ref}(i) - d_Y(j)
\label{eq:1}
\end{equation}
\begin{equation}\small
\Delta f^s (\textsc{Tuffery\_ref}, Y) = f^{s}_\textsc{Tuffery\_ref}(i) - f^{s}_Y(j)
\label{eq:2}
\end{equation}
where the comparison class $Y$ is either a reference class \(Y \in \textsc{Style\_ref}\) or a generated class \(Y \in \textsc{Style\_gen}\), $i$ is the $i$-th vector of \textsc{Tuffery\_ref}, and $j$ is the $j$-th vector of class $Y$.

To evaluate embedding sensitivity to writing style, whether human (\textsc{Style\_ref}) or model-generated (\textsc{Style\_gen}), we compute Pearson correlations, written $r$, between \(\Delta d\) and \(\Delta f^s\) for each stylistic feature \(s\), relative to \textsc{Tuffery\_ref}.

\paragraph{Interpretation.} We interpret the $\Delta d$--$\Delta f^{s}$ correlations as distribution-level associations between dispersion shifts and stylometric shifts across corpora, rather than as causal estimates of rewriting effects (i.e., the indices \(i\) and \(j\) need not refer to paired source--rewrite items). In particular, for \textsc{Style\_gen}, the topic is controlled by construction via rewriting of \textsc{Tuffery\_ref}, so remaining variation is expected to reflect stylistic transfer and generator-specific rewriting behaviour specifically.

%Accordingly, \(r\) quantifies the linear association between changes in embedding dispersion and changes in stylistic features, as defined in Equations~\ref{eq:1} and \ref{eq:2}.

%We thus measure embedding sensitivity to style in each representation: FullD and UMAP reductions to 2D, 3D, and 10D.   

%\begin{equation}
%r^{s}
% = \mathrm{corr}\big(\Delta d, \Delta f^s\big)
%\end{equation}
%where \(\mathrm{corr}\) denotes the Pearson correlation coefficient between pairwise differences. 

\subsection{Embedding Sensitivity to Authorial Style}
\label{ssec:sensihuman}

%To assess separability among the reference corpora with respect to the stylometric features of \citet{terreau2021writing}, we first compared mean normalized feature frequencies across corpora. The aggregated mean values were: $\textsc{Tuffery\_ref}=0.1819$, $\textsc{Proust\_ref}=0.2461$, $\textsc{Celine\_ref}=0.1664$, and $\textsc{Yourcenar\_ref}=0.2070$. An analysis of variance (ANOVA) confirmed significant stylistic differences across these means $(F(3,380)=30.79,\ p=7.73\times 10^{-18})$. Post-hoc Tukey HSD tests revealed that \textsc{Proust\_ref}, \textsc{Celine\_ref}, and \textsc{Yourcenar\_ref} all differed significantly from one another. Moreover, each differed significantly from \textsc{Tuffery\_ref} (all $p<0.05$), except for \textsc{Celine\_ref}, for which the difference from \textsc{Tuffery\_ref} was not statistically detectable at $\alpha=0.05$ with the present sample size ($p=0.298$). Overall, these results confirm that our embedding analyses rest on objectively distinct stylistic profiles.\footnote{Details of the feature-frequency breakdown by reference corpus are available in our GitHub repository under the path: \url{https://anonymous.4open.science/r/style-embedding-sensitivity-BD96/Results/stylo_df_grouped_fr_complet.xlsx}}

%\begin{figure}[h]
%   \centering
%\includegraphics[width=0.9\linewidth]{Figures/corpref.png}
%\caption{Mean frequency across five normalized stylistic features from \citet{terreau2021writing}, by reference corpus.}
%\label{fig:mean_author_ref}
%\end{figure}

To assess embedding sensitivity to human authorial style, we computed the 2D UMAP $\Delta d$--$\Delta f^{s}$ correlations for comparisons between \textsc{Tuffery\_ref} and each author corpus \textsc{Proust\_ref}, \textsc{Celine\_ref}, and \textsc{Yourcenar\_ref}. Figure~\ref{fig:corr_author_ref} reports the sensitivity correlations obtained for each author corpus, broken down by stylistic feature family. For transparency, we report the FullD correlations in Appendix \ref{ssec:fulldsensi} (Figure \ref{fig:corrfulld_author_ref}).\footnote{\label{fn:raw_corr} The full set of raw Pearson correlations for 3D and 10D UMAP, in addition to 2D UMAP and FullD, is available in our GitHub repository.% at: \url{https://anonymous.4open.science/r/style-embedding-sensitivity-BD96/Results/4_sensitivity_correlations_per_corpus_per_dimension.xlsx}
}

For each human author, we observe moderate-to-weak yet significant correlations between embedding-dispersion shifts and stylistic features. \textsc{Proust\_ref} shows broad sensitivity across NER ($r=0.195^*$), Letters ($r=0.163^*$), Structural features ($r=0.136^*$), and TAG ($r=0.127^*$). \textsc{Celine\_ref} concentrates on Letters ($r=0.161^*$), NER ($r=0.132^*$), and Entropy ($r=0.105^*$). Finally, \textsc{Yourcenar\_ref} is dominated by sensitivity to NER ($r=0.212^*$) with minimal Letters and Entropy contributions. 
Overall, these correlations align with the expected authorial stylistic profiles of Proust (syntactic and structural complexity, morphosyntactic richness, proper-names allusions), Céline (spoken register with lexical unpredictability and nonstandard forms), and Yourcenar (strong referential density) described in subsection \ref{ssec:refcorp}.

%The authors were selected to maximize stylistic contrast across four central dimensions in French literature \citep{MagriMourgues2006CorpusStylistique,Brunet2014Lexicometrie,Hough2016OxfordHandbookNames}: structural complexity, morphosyntactic richness, lexical diversity, and referential anchoring. Proust exemplifies structural complexity (long sentences, high average word length) and morphosyntactic richness (diverse POS patterns), with recurrent proper-name allusions (named entities) \citep{brunet1982style}. Céline foregrounds a spoken register (short sentences and words, high pronoun/verb share) alongside lexical diversity and nonstandard forms (higher entropy, distinctive character distributions) \citep{Zabojnikova2006CelineOralPopulaire,Styhre2011CelineHyperbole,SilvestreDeSacy2025Hypersegmentation}.

%Yourcenar sustains balanced prose densely anchored in historical and geographic reference (named-entity density) \citep{Colvin2005BaroqueFictions,Broche2022YourcenarHistoriens}. %Table~\ref{tab:author_styles_nlp} provides an overview of their widely acknowledged stylistic characteristics and the corresponding stylometric features for measurement.

%\footnote{\ref{fig:corrfulld_author}}

%At the level of aggregated features, all correlations were moderate-to-weak but significant ($^*$ indicates $p < 0.01$, after Bonferroni correction, throughout): \textsc{Proust\_ref} ($r = 0.186^*$), \textsc{Celine\_ref} ($r = 0.159^*$), and \textsc{Yourcenar\_ref} ($r = 0.086^*$). 

However, topic variation between \textsc{Style\_ref} and \textsc{Tuffery\_ref} may confound these results. To isolate style from topic, we now examine \textsc{Style\_gen}, where topic is aligned with \textsc{Tuffery\_ref} while style varies through generative rewriting.

\begin{figure*}[t]
    \centering

    \begin{subfigure}[t]{0.30\textwidth}
        \centering
        \includegraphics[width=\linewidth]{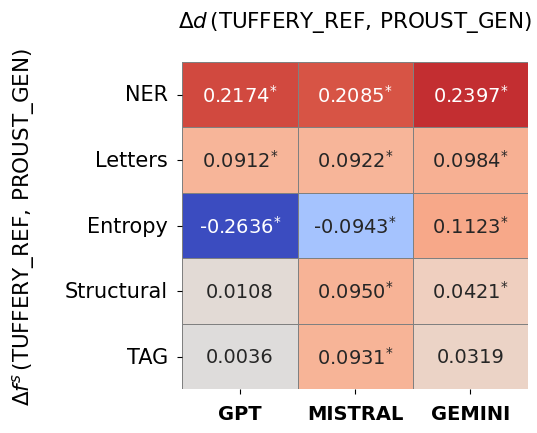}
        \phantomsubcaption\label{fig:perauthorpermodel_a}
        \par\vspace{-1.2em} \makebox[\linewidth][c]{\hspace*{16mm}\small{(a)}}
    \end{subfigure}\hfill
    \begin{subfigure}[t]{0.30\textwidth}
        \centering
        \includegraphics[width=\linewidth]{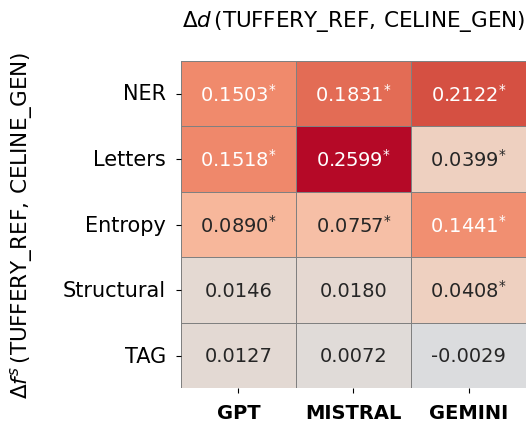}
        \phantomsubcaption\label{fig:perauthorpermodel_b}
       \par\vspace{-1.2em} \makebox[\linewidth][c]{\hspace*{16mm}\small{(b)}}
    \end{subfigure}\hfill
    \begin{subfigure}[t]{0.315\textwidth}
        \centering
        \includegraphics[width=\linewidth]{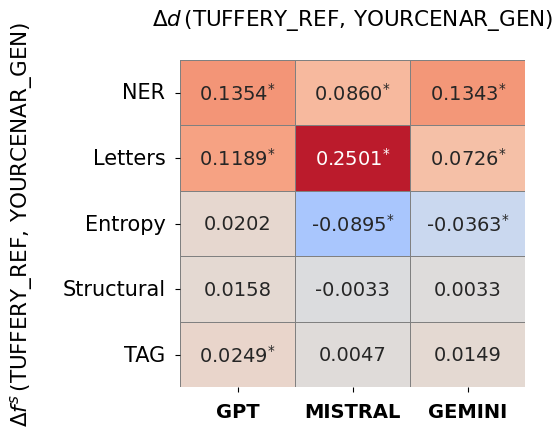}
        \phantomsubcaption\label{fig:perauthorpermodel_c}
        \par\vspace{-1.2em} \makebox[\linewidth][c]{\hspace*{14mm}\small{(c)}}
    \end{subfigure}\hfill
    \begin{subfigure}[t]{0.06\textwidth}
        \centering
        \includegraphics[width=\linewidth]{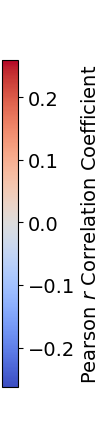}
        \caption*{} % optional (keeps vertical spacing consistent)
    \end{subfigure}

    \vspace{-0.4em}
    \caption{Pearson correlations ($r$) between embedding dispersion shifts in 2D UMAP reduction ($\Delta d$) and stylistic feature shifts ($\Delta f^{s}$) per LLM imitator, comparing
    \textsc{Tuffery\_ref} with (a) \textsc{Proust\_gen}, (b) \textsc{Celine\_gen},
    and (c) \textsc{Yourcenar\_gen}. Asterisks indicate $p<0.01$, with Bonferroni correction.}
    \label{fig:perauthorpermodel}
\end{figure*}

\subsection{Sensitivity After LLM Rewriting}\label{ssec:sensigen}

To assess embedding sensitivity in the context of LLM imitations, we computed the 2D UMAP $\Delta d$--$\Delta f^{s}$ correlations for comparisons between \textsc{Tuffery\_ref} and each imitated-author corpus \textsc{Proust\_gen}, \textsc{Celine\_gen}, and \textsc{Yourcenar\_gen}. Figure~\ref{fig:corr_author_gen} reports the sensitivity correlations for each imitated-author corpus, broken down by stylistic feature family. For transparency again, we report the FullD sensitivity correlations in Appendix \ref{ssec:fulldsensi} (Figure \ref{fig:corrfulld_author_gen}).\footnote{As before, the full set of raw Pearson correlations for all target dimensions is available in our GitHub repository.}
     
%For completeness, Appendix~\ref{ssec:correfull} also reports the FullD correlations (Figure~\ref{fig:corrfulld_author_gen}). 

We observe that \textsc{Proust\_gen} shows highest embedding sensitivity to NER ($r=0.175^{*}$) and Entropy ($r=0.113^{*}$), with weaker effects for Letters and Structural. \textsc{Celine\_gen} concentrates on NER ($r=0.172^{*}$) and Letters ($r=0.143^{*}$). \textsc{Yourcenar\_gen} emphasizes Letters ($r=0.120^{*}$) and NER ($r=0.119^{*}$).

Overall, \textsc{Style\_gen} exhibits moderate stylistic fidelity to the main authorial stylistic dimensions observed in \textsc{Style\_ref}, but these dimensions are attenuated and reweighted. \textsc{Proust\_gen} maintains the broad \textsc{Proust\_ref} stylistic profile in compressed form, with NER only slightly reduced and stronger reductions in Letters, Structural, and TAG. \textsc{Celine\_gen} shows the highest stylistic fidelity among the generated classes,  preserving spoken-style register of \textsc{Celine\_ref} (persistent Letters and weaker Entropy) while overweighting referentiality (NER). \textsc{Yourcenar\_gen} retains the NER-based referential anchoring of \textsc{Yourcenar\_ref}, but more weakly, with partial dilution and a shift toward Letters.

\section{Discussion}
\label{sec:discussion}

\paragraph{LLM Stylistic Fidelity.} Figure~\ref{fig:perauthorpermodel} reports LLM-level sensitivity for the three \textsc{Style\_gen} corpora and reveals author-specific preservation patterns hidden in the pooled results (Figure~\ref{fig:corr_author2D}). 

Figure~\ref{fig:perauthorpermodel_a} shows that \textsc{Mistral} achieves the highest stylistic fidelity for \textsc{Proust\_gen}, preserving Proust’s broad profile as measured in \textsc{Proust\_ref}. (NER, Letters, Structural, TAG). By contrast, Figure~\ref{fig:perauthorpermodel_b} reveals that \textsc{Gpt} achieves higher stylistic fidelity for \textsc{Celine\_gen}, best preserving Céline’s core spoken-style pattern as observed in \textsc{Celine\_ref} (NER, Letters, Entropy). Finally, Figure~\ref{fig:perauthorpermodel_c} shows that \textsc{Gpt} yields greater style fidelity to Yourcenar's defining features in case of \textsc{Yourcenar\_gen} (NER, Letters). %\textcolor{blue}{The weakest model also changes by author (\textsc{Gpt} for Proust, \textsc{Gemini} for Céline, \textsc{Mistral} for Yourcenar), and the failures differ in form: strongest profile compression for Proust with \textsc{Gpt} (loss of Structural and TAG, with negative Entropy), strongest reweighting for Céline with \textsc{Gemini}, and strongest reweighting for Yourcenar with \textsc{Mistral} (weakened NER, over-weighted Letters, and negative Entropy).}

\paragraph{Validation vs Stylistic Fidelity.} These results on LLMs' fidelity to human style only partially align with the style transfer validation results reported in subsection~\ref{ssec:style_transfer_eval}. When evaluating style transfer from \textsc{Style\_ref} to \textsc{Style\_gen} (Table~\ref{tab:style_transfer_validator}), we obtained the highest transfer accuracy for Proust using \textsc{Mistral}, for Céline using \textsc{Mistral} and \textsc{Gemini}, and for Yourcenar using \textsc{Gemini}. However, when assessing stylistic fidelity to the target author profile by LLM (Figure~\ref{fig:perauthorpermodel}), we find that while \textsc{Mistral} still best preserves Proust’s broad stylistic profile, \textsc{Gpt} shows higher stylistic fidelity for Céline than both \textsc{Mistral} and \textsc{Gemini}, and higher stylistic fidelity for Yourcenar than \textsc{Gemini}. 

\paragraph{Explaining Metric Mismatch.} The divergence stems from the two evaluations emphasizing different, though overlapping, stylistic features. Stylistic fidelity measures embeddings preservation of the broad author stylistic profile, while the $3$-$5$-gram-based validator measures surface features, most directly Letters and indirectly NER and TAG.

For Proust, rankings align because \textsc{Mistral} best preserves the Letters-level features that dominate the validator, as well as the target author profile (NER, Letters, Structural, TAG). For Céline, \textsc{Mistral} and \textsc{Gemini} obtain higher style transfer accuracy, consistent with their emphasizing Letters-level surface markers that are highly validator-salient, while \textsc{Gpt} better preserves Céline’s author-defining pattern (NER, Letters, Entropy). For Yourcenar, \textsc{Gemini} similarly achieves higher validator accuracy by matching Letters-level features, whereas \textsc{Gpt} better preserves Yourcenar's defining combination of NER and Letters.

%The same asymmetry also helps interpret the confusion matrix given in Figure~\ref{fig:cm_char_svc}. \textsc{Celine\_gen} is recognized more reliably overall, consistent with \textsc{Mistral} and \textsc{Gemini} (and, more weakly, \textsc{Gpt}) preserving or amplifying surface features that remain discriminative for character $n$-grams. By contrast, \textsc{Proust\_gen} and \textsc{Yourcenar\_gen} are more often confused, consistent with \textsc{Gpt} and \textsc{Gemini} preserving overlapping validator-salient NER and Letters features, while Proust-specific Structural, TAG and Entropy-related differences are only weakly captured. Confusion matrices broken down by imitated author and LLM, reported in Appendix (Figure \ref{fig:perauthorpermodelconfusion}), support these interpretations.

\section{Conclusion}
\label{sec:conclusion}

Our study reveals moderate-to-weak embedding sensitivity to authorial style in French. Comparing literary texts from three authors with their LLM imitations, we find that embeddings broadly reflect author-characteristic stylistic patterns, although the preservation of specific features varies across authors and LLMs. By analyzing stylistic shifts in embedding space, our approach complements style transfer evaluation by tracing how specific features are preserved or altered by generative rewriting.

These findings remain preliminary and require replication across additional languages, broader author sets, and alternative stylistic features. Further analyses should also move beyond aggregated representations to examine more directly how individual embedding architectures encode stylistic features. More broadly, our results suggest that deviations from the human baseline after generative rewriting may help identify LLM-based imitation of authorial style through stylistic transformation patterns. We leave these questions for future work.

%Our study reveals moderate-to-weak embedding sensitivity to targeted stylistic features in French authorial texts. This sensitivity remains detectable after controlled LLM imitation, but is systematically reshaped in feature-specific ways across authors and generators that our analysis can quantify. 

%we aim to evaluate instruction-tuned generative models with explicit style prompts to measure stylistic preservation and determine whether limitations arise from architecture design or from training. 

\section*{Limitations}
\label{sec:limitations}

This study has four main limitations. First, the analyses rely primarily on 2D UMAP because embedding dimensionality varies across models in FullD and because 2D yields the best alignment with both authorial style preservation and FullD purity scores. However, the distortions of authorial features observed in these dimensions require further investigation.

Second, the inventory of stylistic features we consider (NER, Letters, Entropy, Structural, POS Tags) is tailored to the current author sample rather than exhaustive, which may limit coverage of authorial stylistic signals beyond surface features.

Third, the stylistic correlations observed are consistently significant but moderate in magnitude and sometimes limited, calling for broader validation to ensure robustness.

Finally, our study focuses on French literary texts, so generalization to other languages and to non-literary materials in which style is also salient (e.g., correspondence, personal journals, speeches, periodicals), is needed.

%A limitation of our sensitivity analysis is that it relies on correlations computed in a 2D UMAP projection, which can distort distances and thus affect dispersion-based estimates; however, we also observe systematic (and often significant) sensitivity correlations in FullD as well as in 3D and 10D UMAP reductions (reported in the full results file), and we focus on 2D because it yields the sensitivity patterns most consistently aligned with established authorial-characteristic profiles.

\section*{Ethical Considerations}
This work follows the principles of open science, AI transparency, and sustainability, with a strong emphasis on reproducibility and public access to results (lawful use under the EU text-and-data mining exception, Directive 2019/790/EC, Art. 3;\footnote{\url{https://eur-lex.europa.eu/legal-content/FR/TXT/PDF/?uri=CELEX:32019L0790}} in the U.S. context, comparable research use would fall under transformative fair use\footnote{\url{https://www.law.cornell.edu/uscode/text/17/107}}). 

To support open science while respecting copyright regarding the text materials, we release only the 864 LLM-generated rewritings of Tufféry’s texts \textit{in-the-style-of} the three other authors (Proust, Céline, Yourcenar). These constitute transformative stylistic imitations consistent with fair use principles. We do not distribute the 384 original literary texts due to copyright restrictions. We do, however, release the full set of 16,224 vector embeddings ($=$ 1,248 texts $\times$ 13 embedding models) (research-only license; reconstruction or re-identification attempts prohibited). These embeddings are provided for research use only, and we do not evaluate embedding inversion risk in this work. 

To advance AI transparency, our GitHub repository releases all code, the 864 LLM-generated rewritings of Tufféry’s texts, full set of 16,224 embeddings, and analytical results, in particular raw unadjusted Pearson correlations, with documentation for reproducibility. 

To promote sustainability, we used the MTEB leaderboard (Hugging Face) to combine smaller and larger open-source pretrained models in a way that helps limit carbon emissions.

%\section*{Ethical Considerations}

%This work follows the principles of open science, AI transparency, and sustainability, with a strong emphasis on reproducibility and public access to results. To support open science while respecting copyright, we release only the 864 LLM-generated rewritings of Tufféry’s texts “in the style of” the other authors (Proust, Céline, Yourcenar). We do not distribute the 384 original literary texts due to copyright restrictions. We do, however, release the full set of 16,224 vector embeddings (1,248 texts $\times$ 13 embedding models). These embeddings are provided for research use and are not expected to allow reconstruction of the original texts. To advance AI transparency, our dedicated GitHub repository release  all code, unrestricted raw texts, full set of embedding vectors, analytical results, with documentation to enable independent reproducibility. To promote sustainability, we design the framework for net-zero oriented workflows and balance resource use by combining smaller open-source pretrained models with larger ones, reducing carbon footprint while maintaining models diversity and methodological rigor.

\section*{Acknowledgements}

We thank Stéphane Tufféry for authorizing us to use the material from his book \textit{Le style mode d'emploi} in our study, as well as two anonymous reviewers for helpful comments and feedback. This work was supported by the programs THEMIS (grant agreements n°DOS022279400 and n°DOS022279500) and TRUSTEDNEWS (ANR-25-ASM2-0003). EZ acknowledges Infopro Digital for supporting her PhD research, alongside her work. 

\section*{Declaration of Contribution}

BI conceptualized the research problem and designed the experiment. LS and AB managed the data collection and generation processes. LS was responsible for coding and testing the selected generation models, while BI managed the style evaluation models. BI analyzed and discussed the results, with LS and JGG. BI wrote the paper, which was read and revised collaboratively by all authors. Correspondence: benjamin.icard@lip6.fr.

%1248$\times$13=16224

%\section*{Supplementary Materials}
%Embedding models used in the experiments included the RoBERTa-based models \texttt{xlm-roberta-large},\footnote{\url{https://huggingface.co/FacebookAI/xlm-roberta-large}} the 1024-dimensional models \texttt{xlm-roberta-large}\footnote{\url{https://docs.mistral.ai/capabilities/embeddings/}} by Mistral, \texttt{distilbert-base-uncased},\footnote{\url{https://huggingface.co/distilbert/distilbert-base-uncased}} the 1536-dimensional model \texttt{text-embedding-3-small}\footnote{\url{https://platform.openai.com/docs/guides/embeddings}} by OpenAI, \texttt{voyage-2}\footnote{\url{https://docs.voyageai.com/docs/embeddings}} by Voyage, \texttt{multilingual-e5-large},\footnote{\url{https://huggingface.co/intfloat/multilingual-e5-large}} and \texttt{all-roberta-large-v1}.\footnote{\url{https://huggingface.co/sentence-transformers/all-roberta-large-v1}} Smaller embedding models included the 768-dimensional models \texttt{e5-base-v2},\footnote{\url{https://huggingface.co/intfloat/e5-base-v2}} the SBERT model \texttt{sentence-camembert-base},\footnote{\url{https://huggingface.co/dangvantuan/sentence-camembert-base}} \texttt{text-embedding-004},\footnote{\url{https://cloud.google.com/vertex-ai/generative-ai/docs/embeddings/get-text-embeddings}} the multilingual model \texttt{paraphrase-multilingual-mpnet-base-v2},\footnote{\url{https://huggingface.co/sentence-transformers/paraphrase-multilingual-mpnet-base-v2}} \texttt{all-MiniLM-L12-v2}.\footnote{\url{https://huggingface.co/sentence-transformers/all-MiniLM-L12-v2}} We also included \texttt{solon-embeddings-large-0.1}\footnote{\url{https://huggingface.co/OrdalieTech/solon-embeddings-large-0.1}} (1024D) by Solon as one of the best performing French embedding models according to the MTEB leaderboard on HuggingFace\footnote{\url{https://huggingface.co/spaces/mteb/leaderboard}} at the time of submission. 

%All the code, data (unrestricted raw texts, embeddings of restricted and unrestricted texts), and analytical results are available at: \url{https://anonymous.4open.science/r/style-embedding-sensitivity-BD96}

%\section*{Acknowledgments}

%We thank Stéphane Tufféry for giving us access to his book \textit{Le style mode d'emploi}

%\section*{Declaration of Contribution}

%\nocite{*}
%\section*{References}\label{sec:reference}

% ACL bibliography style:
%\bibliographystyle{acl_natbib}
\bibliography{custom.bib} % <-- rename to your .bib file name if needed

\appendix

\begin{table*}[t]
\centering
\small
\resizebox{\linewidth}{!}{%
\begin{tabular}{llccc||c}
\hline
\multicolumn{2}{l}{\textbf{Validation Set}}  &
\textsc{Proust\_ref} & \textsc{Celine\_ref} & \textsc{Yourcenar\_ref} & \textsc{Style\_ref} (held-out 20\%) \\ \hline
\multirow{2}{*}{\shortstack[l]{Corpus-Level}} & Macro-F1
& --- & --- & --- & 0.825 \\
& Accuracy
& 0.737 & 0.789 & 0.950 & \textbf{0.828} \\

\hline
\end{tabular}}
\vspace{1em}

\resizebox{\linewidth}{!}{%
\begin{tabular}{llccc||c}
\hline
\multicolumn{2}{l}{\textbf{Test Set}} &
\textsc{Proust\_gen} & \textsc{Celine\_gen} & \textsc{Yourcenar\_gen} & \textsc{Style\_gen} (100\%) \\ \hline

\multirow{2}{*}{\shortstack[l]{Corpus-Level}} & Macro-F1
& --- & --- & --- & 0.528 \\
& Transfer Accuracy
& 0.403 & 0.438 & 0.760 & \textbf{0.534} \\ \hline

\multirow{3}{*}{\shortstack[l]{Per-Class}}
& \quad \textit{when \textsc{GPT} is used}
& 0.260 & 0.385 & \textcolor{blue}{0.854} & 0.500 \\ 
& \quad \textit{when \textsc{Mistral} is used}
& \textcolor{blue}{0.604} & \textcolor{blue}{0.646} & 0.625 & \textcolor{blue}{0.625} \\ 
& \quad \textit{when \textsc{Gemini} is used}
& 0.344 & 0.281 & 0.802 & 0.476 \\ 
\hline
\end{tabular}}
\caption{Style transfer results with Function words Frequencies + \texttt{LinearSVC}. We report corpus-level performances on \textsc{Style\_ref} (held-out 20\%) and \textsc{Style\_gen} (100\%), and per-class transfer accuracy for each target author label. We indicate corpus-level accuracy in bold, and the highest per-class transfer accuracy across LLMs in blue.}
\label{tab:style_transfer_validator_function}
\end{table*}

\section{Appendix}

\subsection{Style Transfer Evaluation Using Function Words}
\label{ssec:styletransfunct} 

Table~\ref{tab:style_transfer_validator_function} reports style transfer performances obtained on the corpus with the validator using function word frequencies in combination to \texttt{LinearSVC}.

At the corpus level, this validator yields weaker but overall consistent corpus-level results compared to $3$-$5$-gram-based validator, with lower performance on both \textsc{Style\_ref} ($0.828$ vs. $0.966$ accuracy; $0.825$ vs. $0.965$ Macro-F1) and \textsc{Style\_gen} ($0.534$ vs. $0.664$ transfer accuracy; $0.528$ vs. $0.669$ Macro-F1), while preserving the same contrast between validation on reference texts and weaker transfer on generated texts. 

Across reference authors (\textsc{Proust\_ref}, \textsc{Celine\_ref}, \textsc{Yourcenar\_ref}), both validators agree in assigning the highest accuracy to Yourcenar ($0.950$ vs. $1.000$), although the function-word-based model is notably less accurate for Proust and Céline.

Across imitated author (\textsc{Proust\_gen}, \textsc{Celine\_gen}, \textsc{Yourcenar\_gen}) and LLM combinations, the picture is partly stable and partly shifted: \textsc{Mistral} remains the best model for \textsc{Proust}, and also remains best for \textsc{Celine}, though without the earlier tie with \textsc{Gemini}. For \textsc{Yourcenar}, however, the best system shifts from \textsc{Gemini} under the character \textit{n}-gram validator to \textsc{GPT} under the function-word-based validator, while \textsc{Gemini} remains strong.

\subsection{Confusion Matrix for the $3$-$5$-Gram- Based Validator}
\label{ssec:appendix_confusions}

\begin{figure}[htbp]
   \centering
\includegraphics[width=\linewidth]{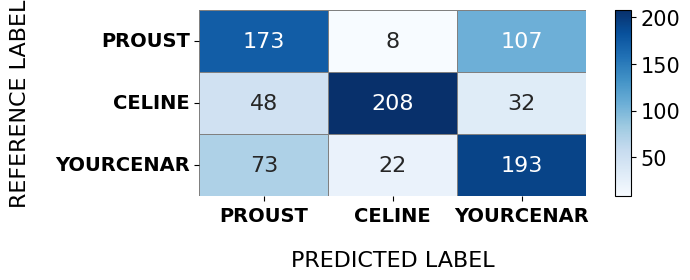}
\caption{Confusion matrix for the \textsc{Style\_gen} evaluation of the TF-IDF character $3$-$5$-gram + \texttt{LinearSVC} classifier trained on \textsc{Style\_ref}, shown by imitated authorial style. Rows correspond to true author labels and columns to predicted labels.}
\label{fig:cm_char_svc}
\end{figure}

Figure~\ref{fig:cm_char_svc} complements the aggregate performance results given in Table \ref{tab:style_transfer_validator} for the $3$-$5$-gram-based validator, showing the class-specific error structure on \textsc{Style\_gen}. The dominant errors are mutual confusions between \textsc{Proust\_gen} and \textsc{Yourcenar\_gen} (107 \textsc{Proust\_gen} instances predicted as \textsc{Yourcenar\_ref}, and 73 \textsc{Yourcenar\_gen} instances predicted as \textsc{Proust\_ref}), whereas misclassifications into \textsc{Celine\_ref} are comparatively rare for the other two styles (8 and 22).

%\subsection{Projection Plot for Best 2D UMAP Model}

% \begin{figure*}[t]
%     \centering
% \captionsetup[subfigure]{justification=centering,format=plain}
%     \begin{subfigure}[t]{0.32\textwidth}
%         \centering
%         \includegraphics[width=\linewidth]{Figures/gpt_confusion.png}
%         %\caption{\textsc{Gpt}.}
%         \par\vspace{-0.2em} \makebox[\linewidth][c]{\hspace*{8mm}\small{(a)}}
%     \end{subfigure}\hfill
%     \begin{subfigure}[t]{0.32\textwidth}
%         \centering
%         \includegraphics[width=\linewidth]{Figures/mistral_confusion.png}
%         %\caption{\textsc{Mistral}.}
%         \phantomsubcaption\label{fig:perauthorconfusion_b}
%        \par\vspace{-1.5em} \makebox[\linewidth][c]{\hspace*{8mm}\small{(b)}}
%     \end{subfigure}\hfill
%     \begin{subfigure}[t]{0.32\textwidth}
%         \centering
%         \includegraphics[width=\linewidth]{Figures/gemini_confusion.png}
%         %\caption{\textsc{Gemini}.}
%         \phantomsubcaption\label{fig:perauthorconfusion_c}
%         \par\vspace{-1.5em} \makebox[\linewidth][c]{\hspace*{9mm}\small{(c)}}
%     \end{subfigure}\hfill
%
% \vspace{-0.5em}
% \caption{Confusion matrices for \textsc{Style_gen} style-transfer validation, broken down by LLM generator: (a) when \textsc{Gpt} is used, (b) when \textsc{Mistral} is used, and (c) when \textsc{Gemini} is used.}
%   \label{fig:perauthorpermodelconfusion}
% \end{figure*}

\subsection{FullD Sensitivity Correlations to Style}
\label{ssec:fulldsensi}

\begin{figure*}[!h]
    \centering
    \begin{subfigure}[t]{0.46\textwidth}
        \centering
        \includegraphics[width=\linewidth]{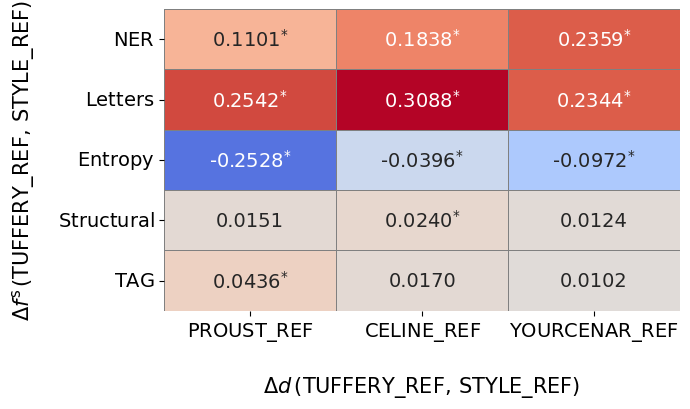}
       
        \phantomsubcaption\label{fig:corrfulld_author_ref}
        \par\vspace{-0.2em} \makebox[\linewidth][c]{\hspace*{17mm}\small{(a)}}
        
        %\caption{Correlations ($r$) between embedding dispersion shifts ($\Delta d$) and stylistic feature shifts ($\Delta f^s$) for \textsc{Style\_ref} three corpora relative to \textsc{Tuffery\_ref}. Asterisks indicate $p<0.01$, with Bonferroni correction.}

        \label{fig:corrfulld_author_reffig}
        
    \end{subfigure}
    \hfill
    \begin{subfigure}[t]{0.46\textwidth}
        \centering
        \includegraphics[width=\linewidth]{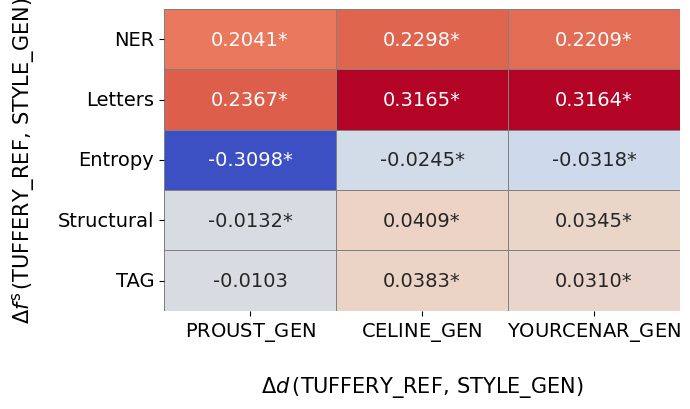}
        \phantomsubcaption\label{fig:corrfulld_author_gen}
        \par\vspace{-1.3em} \makebox[\linewidth][c]{\hspace*{17mm}\small{(b)}}
        \label{fig:corrfulld_author_genfig}
    \end{subfigure}

    \vspace{-1em}
    \caption{Pearson correlations ($r$) between embedding dispersion shifts in FullD ($\Delta d$) and stylistic feature shifts ($\Delta f^s$) for each author-labeled corpus, comparing \textsc{Tuffery\_ref} with (a) the three human-authored corpora in \textsc{Style\_ref}, and (b) the three style-imitated corpora in \textsc{Style\_gen}. Asterisks indicate $p<0.01$ after Bonferroni correction.}
        \label{fig:corrfulld_authorfig}
\end{figure*}

Figure~\ref{fig:corrfulld_authorfig} reports sensitivity correlations between changes in stylistic features and changes in FullD embedding dispersion across authors, comparing \textsc{Tuffery\_ref} with both human-written corpora (\textsc{Style\_ref}) and style-imitated (\textsc{Style\_gen}) corpora. 

Compared to 2D UMAP (Figure~\ref{fig:corr_author_ref}), Figure~\ref{fig:corrfulld_author_ref} shows that many feature-wise correlations remain significant in FullD and some are actually stronger than in 2D UMAP, but the resulting profiles are less faithful to the authorial characteristics of the human authors. For \textsc{Proust\_ref}, the expected broad profile becomes distorted: Letters dominates, while Structural is nearly absent and NER is reduced. For \textsc{Celine\_ref}, the expected concentration on Letters, NER, and Entropy is weakened by a reversal of Entropy and the appearance of Structural. For \textsc{Yourcenar\_ref}, the expected referential dominance of NER becomes less distinctive because Letters rises to a comparable level and Entropy turns negative. Overall, compared with 2D UMAP, the FullD correlations differentiate the three human authors less clearly because they overweight shared features, especially Letters and, to a lesser extent, NER, while weakening the author-specific balance of stylistic features.

Compared to 2D UMAP (Figure~\ref{fig:corr_author_gen}), Figure~\ref{fig:corrfulld_author_genfig} shows that these distortions become even stronger after LLM rewriting. In \textsc{Proust\_gen}, the profile narrows to NER and Letters, while Entropy becomes strongly negative and Structural and TAG largely disappear. In \textsc{Celine\_gen}, Letters and NER remain prominent, but Entropy again turns negative and Structural and TAG become more visible than expected for Céline. In \textsc{Yourcenar\_gen}, the profile becomes flatter and less specific than in 2D, with strong Letters and NER accompanied by positive Structural and TAG and negative Entropy. 

FullD therefore preserves authorial characteristic features less well than 2D UMAP, both for human-authored texts and after LLM rewriting, because the more selective feature-level preservation observed in 2D is replaced by a more generic pattern shared across corpora. Our GitHub repository shows similar distortions with the 3D and 10D UMAP reductions.

\end{document}